\documentclass[letterpaper]{article} \usepackage{aaai2027}  \nocopyright \usepackage[hyphens]{url}  \usepackage{graphicx}    \usepackage{natbib}  \usepackage{caption}   \usepackage{booktabs}
\usepackage{array} \usepackage{xcolor} 
\graphicspath{{Figures/}}

\newcommand{\swatch}[1]{\textcolor{#1}{\rule{1.6ex}{1.6ex}}}
\newcommand{\predgtlegend}{  \small\swatch{red}~Surrogate prediction\quad\swatch{gray}~Flow-analysis ground truth}

\title{From AI Surrogate to Engineering Tool: Toward Production Use of Fiber-Orientation Prediction in Crash CAE}
\author{
Osamu Ito\textsuperscript{\rm 1,2}\corresponding,
Yoshikazu Nakagawa\textsuperscript{\rm 1},
Tomoaki Tatsukawa\textsuperscript{\rm 2}
}
\affiliations{
\textsuperscript{\rm 1}Honda R\&D Co., Ltd., 4630 Shimotakanezawa, Haga-machi, Haga-gun, Tochigi 321-3393, Japan\\
\textsuperscript{\rm 2}Tokyo University of Science, 6-3-1 Niijuku, Katsushika-ku, Tokyo 125-8585, Japan\\
osamu\_ito@jp.honda
}

\begin{document}
\maketitle

\begin{abstract}
Crash simulations of glass-fiber-reinforced plastic parts are highly sensitive to the fiber-orientation field supplied to the material model. In conventional workflows, that field comes from resin-flow analysis, which requires molding-specialist support and takes about 25 hours of work per part. We previously proposed a surrogate model that addresses this bottleneck by predicting the orientation field in about 3 minutes. This paper reports the implementation, validation, and operational design that turned the model into an engineering tool for Honda design workflows. We present staged validation from 2D plates to production parts, evaluation against a predefined engineering angular-error criterion, and the expert review that approved design use in March 2025; the system has since been transferred to the crash-safety group and is in initial, limited use in directed fracture-prediction scenarios. The developed system combines injection-gate setting support for cases in which gates are unknown ($90\%$ gate-count accuracy, mean position error $11\%$ of flow length), voxelization, surrogate inference, automatic mapping to the crash CAE model, and report generation in a single GUI. The complete orientation-preparation workflow takes about 0.1 hours, a 99.6\% reduction from the conventional 25-hour flow-analysis route, and can be run directly by crash CAE engineers without molding-specialist support. As evidence of practical impact, component-level crash analysis using the predicted orientation field reduced the absolute peak-force error from 19.0\% with an isotropic baseline to 11.4\%, while confining predicted fracture to the test-observed region. The deployment lesson is that a surrogate model alone is not enough for practical adoption: gate-setting support, automatic data conversion, and shareable reporting are what make the model usable as a design tool.
\end{abstract}

\section{Introduction}

\begin{figure}[t]
\centering
\includegraphics[width=\columnwidth]{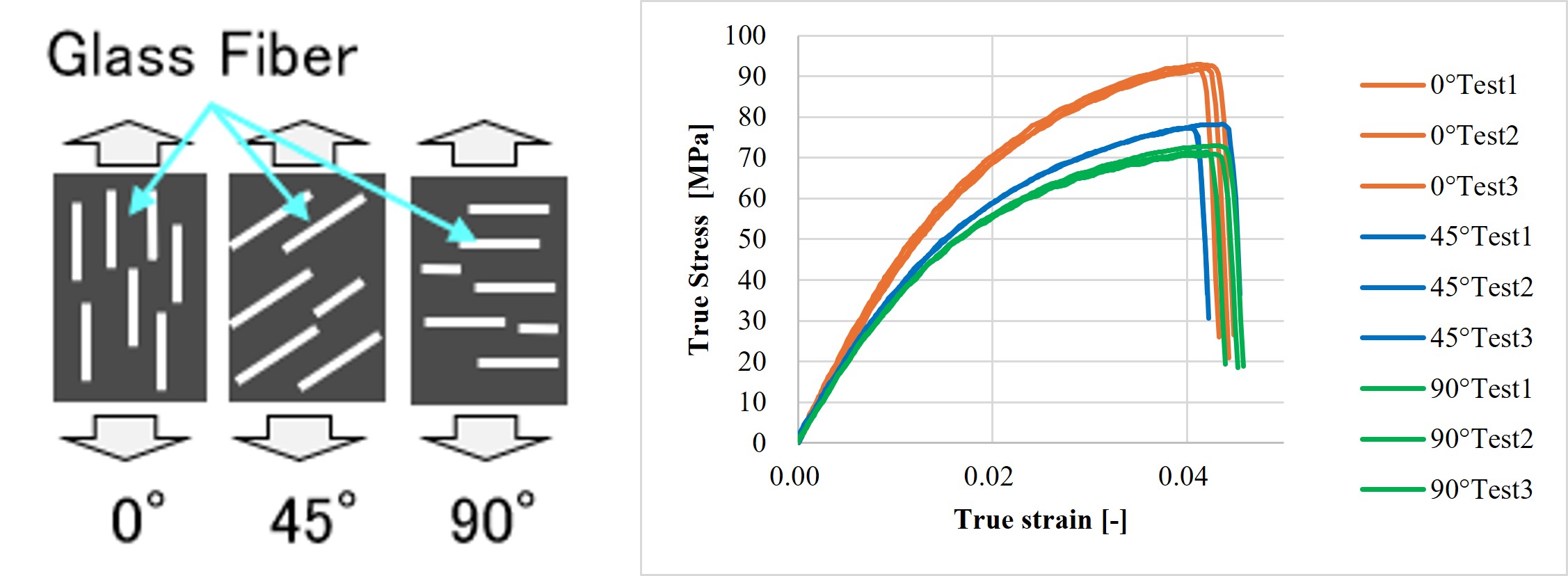}
\caption{Effect of fiber orientation on PP-GF tensile response. Specimens cut at $0^\circ$, $45^\circ$, and $90^\circ$ to the fiber direction (left) show distinct true stress--strain responses (right), with peak stress ranging from about 70 to 92\,MPa.}
\label{fig:aniso}
\end{figure}

Automotive structures are increasingly replacing selected steel components with fiber-reinforced plastics to reduce weight and increase design freedom. In short-glass-fiber-reinforced polypropylene (PP-GF), stiffness, strength, and fracture are strongly anisotropic because they depend on the local fiber direction set during injection molding. This effect is large enough to matter in crash analysis: specimens cut at $0^\circ$, $45^\circ$, and $90^\circ$ to the fiber direction differ by roughly $25\%$ in peak stress (Figure~\ref{fig:aniso}). Because the load response depends on the fiber direction, a crash model that ignores orientation can mispredict deformation and fracture.

Orientation-aware crash CAE is therefore both necessary and technically feasible. In prior work, impact-bending tests on specimens cut from production brackets showed that an anisotropic crash model supplied with flow-analysis orientation reproduces location-dependent load--displacement response and fracture, whereas an isotropic model does not \citep{ito2026jsme,fertschej2016impetus}. The practical barrier is the cost of obtaining the orientation field. Generating it with a resin-flow solver such as 3D~TIMON and mapping it to an LS-DYNA anisotropic material card \citep{timon2026,hallquist2006lsdyna} requires molding-specialist work and takes about 25 hours per part. Vehicle programs iterate geometry on a much shorter cadence, so fiber-reinforced plastic parts are often still modeled as isotropic in everyday practice, with the resulting error accepted as a workflow compromise.

To remove this bottleneck, we previously developed a machine-learning surrogate trained on paired flow analyses; for a new part, it predicts the fiber-orientation field in about 3 minutes \citep{ito2026jsme}. However, an accurate predictor is not yet an engineering tool. Moving toward operational use revealed requirements that the model alone did not satisfy: an interface that crash CAE engineers can operate without molding expertise, support for specifying injection gates when suppliers have not disclosed them, automatic mapping to the crash model and material cards, and shareable reports for design discussion. Learned predictors of physical fields are now common \citep{guo2016flowcnn,thuerey2020airfoil,vinuesa2022cfdml}, typically built on image-translation or U-Net backbones \citep{isola2017pix2pix,ronneberger2015unet}, and crash accuracy for fiber-reinforced plastics has been improved by passing moldfilling-derived anisotropy and failure criteria to crash solvers \citep{nutini2010anisotropy,nutini2019failure}. What is less well documented is how such a predictor becomes a system suitable for routine use; that process is the subject of this paper.

Accordingly, this paper does not propose a new surrogate model. Instead, it reports how a previously published model was developed into a system that crash CAE engineers can use in design work, and what we learned while putting it into practical use. The paper covers:
{\setlength{\itemsep}{0pt}\setlength{\parskip}{0pt}\setlength{\parsep}{0pt}
\begin{itemize}
\item the application, requirements, and prior surrogate (Sections~2--3);
\item staged validation and the annular-part coverage gap (Section~4);
\item practical impact in a pedestrian-impact case (Section~5);
\item integrated tooling for gate prediction, mapping, reporting, and GUI operation (Section~6); and
\item development cost, expert review, rollout status, and lessons for similar projects (Section~7).
\end{itemize}}

The distinction from the prior publication is as follows. The earlier paper in the Transactions of the JSME reported the surrogate architecture and component-level crash validation \citep{ito2026jsme}. In contrast, this IAAI paper treats that model as the technical foundation and reports the operationalization process around it: gate-setting support, automatic mapping, report generation and the integrated GUI, acceptance validation, expert review and transfer, and maintenance practice. Selected figures from the prior paper are reused here only as cited background, not as the contribution of this paper.

\section{Application Description}

Pedestrian-protection testing, a major part of crash-safety assessment, increasingly requires accurate prediction of deformation and fracture in bumper-area components, where polymer parts are widely used. For example, the newer aPLI legform adds a simplified upper-body mass and increases the impactor mass from 13 to 25\,kg relative to its FlexPLI predecessor \citep{humanetics2022apli}. As the loading becomes more severe, conservative design margins are a poor substitute for reproducing the deformation itself. The CAE model must capture the part's direction-dependent load--deformation response, which requires the local fiber orientation to be represented in the material model.

The intended users are crash CAE engineers. They evaluate, before the design geometry is frozen, whether a fiber-reinforced part satisfies crash and pedestrian-protection targets. Accurate assessment requires the fiber-orientation field inside the part---an input normally obtained from resin-flow analysis. In routine design loops, however, there is often no time to run flow analysis for every shape update, so the part is analyzed as isotropic. The surrogate must therefore generate the required orientation information from inputs the engineer already has: a CAD model and a candidate gate layout (the material grade is fixed to PP-GF20, the single grade the system covers). The required output is not a visually faithful flow-analysis image, but core- and skin-layer orientation vectors that the crash material card can consume directly.

The system is designed to replace only the repeated resin-flow-analysis step, not the surrounding engineering process. The workflow still follows the existing chain from design input (CAD, gate layout) through voxelization, core/skin orientation prediction, material-card mapping, and crash judgment. This drop-in property is essential for adoption because it avoids retraining the CAE team or changing downstream tools. It also changes the iteration cadence: because the orientation field is available in minutes rather than a day, engineers can generate it for each candidate design and use it in crash analysis within the same design iteration.

\begin{figure}[!htbp]
\centering
\includegraphics[width=0.74\columnwidth]{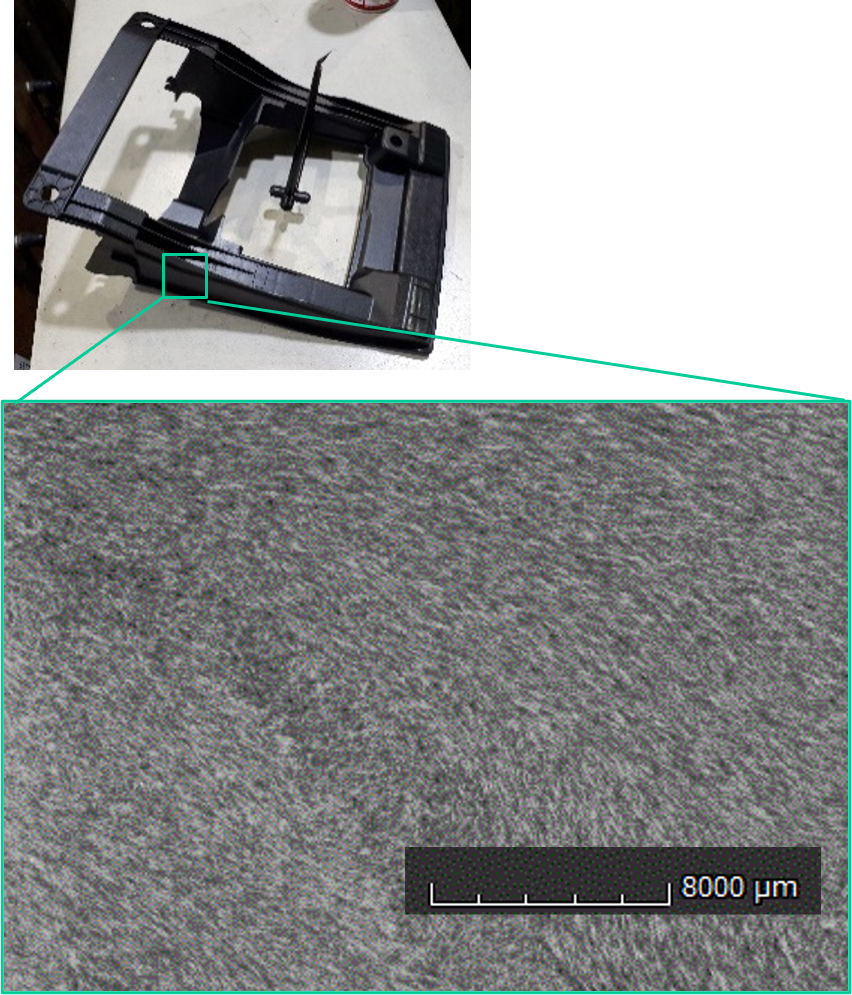}
\caption{Target PP-GF bracket and X-ray CT observation region. The enlarged CT image shows the local orientation texture formed by short glass fibers inside the part; scale bar: $8000\,\mu\mathrm{m}$.}
\label{fig:part}
\end{figure}

The application target is a PP-GF20 bracket. Figure~\ref{fig:part} shows the part and an X-ray CT observation region, illustrating that short glass fibers form locally oriented texture inside the part; the representative fiber length measured by X-ray CT is 0.5\,mm. The part was validated in a pedestrian-impact scenario using a 4.5\,kg impactor at 4.5\,m/s. Sections~4 and~5 show how the surrogate was validated for this part and what it changed in practice.

\FloatBarrier

\paragraph{Requirements.} The engineering acceptance criteria were defined before development. They are not generic ML metrics but criteria for routine crash CAE use: orientation prediction within 5 minutes, so the field can be regenerated inside a single design iteration; an impactor peak-load error within $20\%$ of the component test when the predicted field drives the crash model; predicted fiber direction within $45^\circ$ of the reference in regions where orientation affects the material response; and output mappable into the anisotropic crash material card with no manual repair. A further constraint shaped the tooling: adoption must add no time to the build schedule of the full-vehicle CAE model it feeds.

\section{AI Technology: A Voxel-Based 3D Conditional GAN}

This section summarizes the previously developed surrogate only to the extent needed for the validation and tooling reported here: how part geometry and gate conditions are encoded, and how the model returns the core- and skin-layer orientation information crash CAE requires.

\paragraph{Representation.} The surrogate input places part geometry and gate conditions on one three-dimensional grid: the geometry is rasterized into a uniform $200\times200\times200$ voxel grid whose edge length, at most about 2\,mm, matches the downstream crash shell mesh, and each occupied voxel carries the Euclidean distance to each gate. The distance fields hand the network the inflow boundary condition directly---which inflow source governs each location and where flow fronts are likely to meet---following the observation that distance-to-boundary features are predictive in learned flow approximation \citep{guo2016flowcnn}. Voxelization decouples the surrogate from the flow solver's finite-element mesh and gives every part a fixed-size input. The output has six channels: core- and skin-layer orientation vectors with the corresponding orientation degrees, predicted separately because injection molding creates the two layers physically---and the crash material card consumes both.

\paragraph{Network.} The model lifts the Pix2Pix paired-translation idea \citep{isola2017pix2pix} to 3D (Figure~\ref{fig:network}, Appendix~A): a 3D U-Net generator \citep{ronneberger2015unet} with skip connections and a local ($1\times1\times1$) PatchGAN discriminator that enforces per-voxel orientation consistency. Training combines a binary cross-entropy adversarial loss with an MSE reconstruction loss weighted by $\lambda{=}100$. The adversarial term was included because orientation fields contain sharp local structure---weld lines, stagnation points, shear layers along edges---that a pure-MSE regressor tends to blur toward the mean; the discriminator is intended to push the generator toward locally crisp, physically plausible patterns. Separate one-gate and two-gate models are trained because weld-line behavior differs between them.

\paragraph{Manufacturability-constrained data generation.} The decisive design choice is what the model is trained \emph{on}. Gate positions are not sampled freely: a survey of 46 production parts yielded the design rules molding engineers actually follow---the flow ratio $FR{=}L/t$ (flow length over wall thickness) must stay below about 130; a part volume of about 200{,}000\,mm$^3$ separates side-gate from direct-gate designs (with exceptions for lattice structures and for surfaces too narrow to seat a direct gate); and gates are placed toward the part center to minimize flow distance. Applying these rules, 20 base shapes were crossed with 100 manufacturable gate patterns each, for both gate counts, yielding 4{,}000 paired flow analyses in total under injection speeds of 10--28\,mm/s and a pressure limit of 100\,MPa. Sampling gates at random would flood training with configurations no mold designer would build, forcing the model to spend capacity on spurious flow features; constraining the data to manufacturable settings likely contributed to the achieved accuracy. The application-specific AI design choice is thus the use of manufacturing constraints to shape the training distribution, rather than a new network architecture.

\section{Staged Validation: From Plates to a Production Part}

Committing to the production system meant 4{,}000 flow solves and on the order of a thousand hours of training, so we de-risked the approach in stages on controlled geometries first; the same stages now serve as the application's trust argument. The order was deliberate: each stage raised data-generation and training cost in step with confidence, so a failure of the basic idea would have surfaced at the cheapest possible point. Stages 1--2 report mean squared error (MSE) of the predicted field against a 0.03 target; Stage 3 reports regional angular error in degrees, the quantity the material card is most sensitive to. Throughout, ``ground truth'' denotes the resin-flow solution---the reference the surrogate is trained to match, not absolute physical truth, a distinction that matters once below. Table~\ref{tab:metrics} summarizes all stages.

\subsection{Stages 1--2: Controlled Plate Studies}

The first stage confirmed, on a $50\times150\times2.5$\,mm three-gate plate with a triangular cutout (22{,}044 image-based training pairs from 22 cutout-and-gate families), that a conditional GAN can learn resin-flow-induced orientation: across 4{,}845 unseen cases the maximum MSE was 0.0175, well below the 0.03 target. This stage also exposed an unexpected robustness property (Figure~\ref{fig:perturb}). In the worst case the flow-analysis reference gave a transverse orientation that additional flow-analysis runs, each with one gate moved by 3\,mm, did not reproduce---an isolated unstable solver result that the surrogate, trained over many neighboring conditions, smoothed toward its neighbors. This does not make the surrogate generally more correct than the solver, but it shows why the system must surface local high-error regions, and analysts need to see where.

The second stage introduced a controlled 3D T-plate: a base plate plus a $35\times20\times2.5$\,mm rib, in seven shape families with hole variants. Though simplified, it captures three effects that matter before production parts: the rib drives out-of-plane flow, holes create local disturbances and weld lines, and three gates produce interacting flow fronts. Because a rendered 2D image cannot represent flow over the rib, the input was changed to per-voxel gate-distance fields with orientation-vector targets---the representation later used for production parts. On 1{,}050 unseen-\emph{shape} cases, inference took about one minute each, with MSE from 0.0004 to 0.0248 (average 0.0029; Figure~\ref{fig:tplate}), all below target. The largest residuals localized predictably at sharp-cornered holes and their short weld lines rather than spreading across the field.

\subsection{Stage 3: Production-Part Holdout}

\begin{figure}[t]
\centering
\includegraphics[width=0.87\columnwidth]{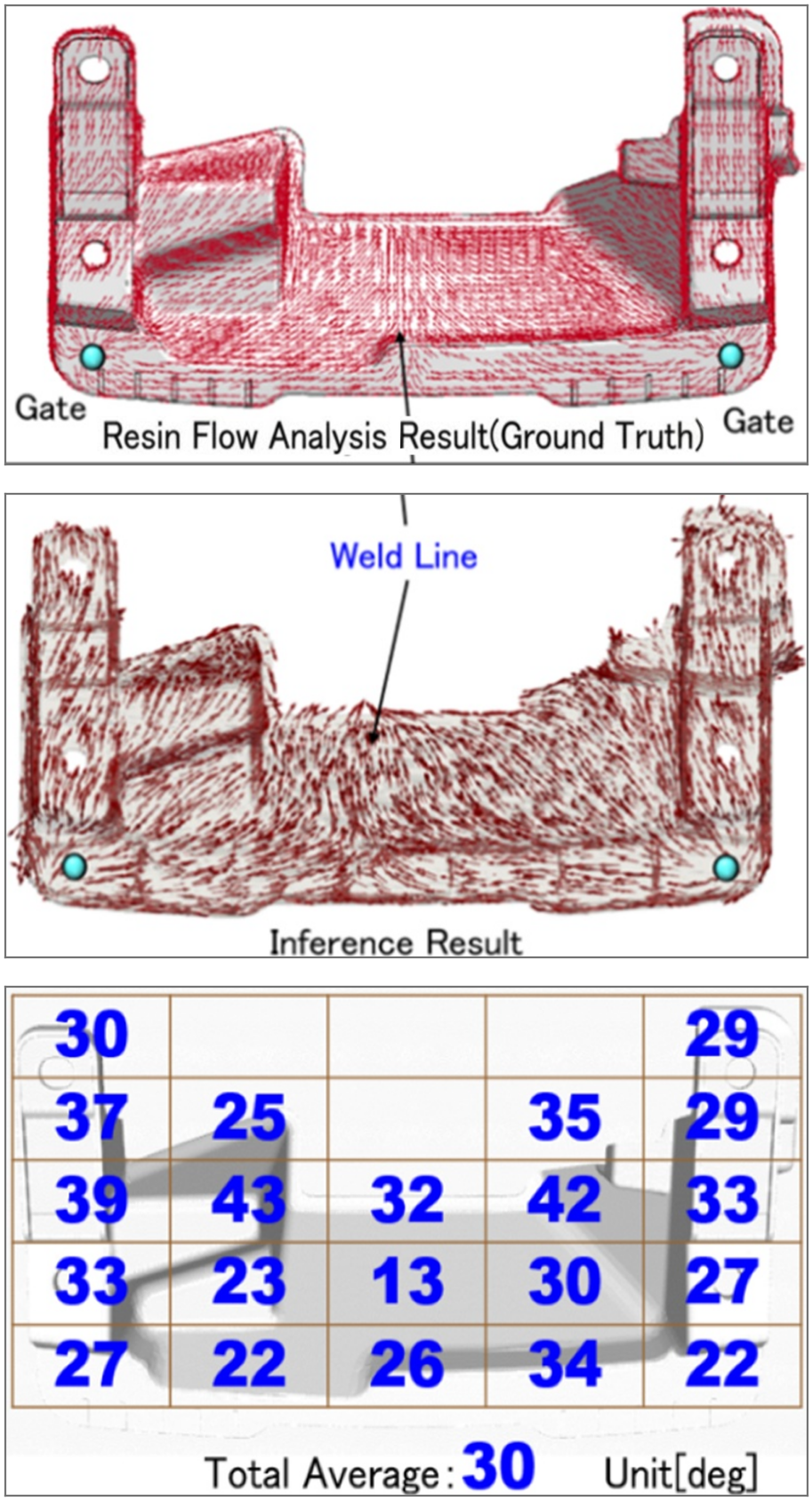}
\caption{Stage 3, two-gate production part: ground-truth flow field (top), surrogate prediction reproducing the weld line (middle), and the $5\times5$ regional angular-error map in degrees (bottom; mean $30^\circ$, within the $45^\circ$ anisotropy-preserving bound).}
\label{fig:2gate}
\end{figure}

The production model of Section~3 was then validated on held-out parts (one per gate count). Accuracy is the per-voxel angular error between predicted and reference vectors, averaged over a $5\times5$ region grid to reveal spatial bias. The single-gate part reproduced the radial spread at a mean regional error of $22^\circ$; the two-gate part reproduced the dominant field and the merge-line weld---the hardest feature---at $30^\circ$ (Figure~\ref{fig:2gate}). The $45^\circ$ bound is the predefined engineering acceptance criterion, chosen because laminate in-plane stiffness transforms with $\cos^4\theta$/$\sin^4\theta$ terms that cross at $45^\circ$, where fiber-direction and transverse contributions become equal; below it, the predicted anisotropy still informs rather than corrupts the response \citep{ito2026jsme}. Both holdouts, at $22^\circ$ and $30^\circ$, meet this criterion.

Validation also surfaced a coverage gap. The first annular (ring-like) holdout missed the $20\%$ peak-load target ($+21\%$): the merge-region field was wrong because only 2 of the 20 training shapes were annular---too few to learn closed-loop merging. A dedicated annular model, same recipe, restored that field (regional mean $22^\circ$) and cut the load error to $+11.4\%$ (the value used in Section~5).

\section{Application Use and Practical Impact}

The first effect is preparation time. Inference itself takes about 3 minutes; with preprocessing, mapping, export, and overhead, the end-to-end workflow takes about 0.1 hours versus roughly 25 hours for the flow-analysis route---a 99.6\% reduction; gate setting and the crash solve itself lie outside both routes---making orientation-aware modeling repeatable within each design iteration. It also changes \emph{who} runs it: the flow route needs a molding specialist, whereas crash engineers operate the workflow directly, removing a cross-team dependency that used to add days.

\begin{figure}[t]
\centering
\includegraphics[width=0.96\columnwidth]{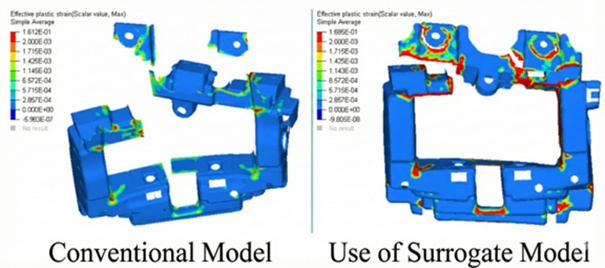}\\[1pt]
{\small (a) Effective plastic strain at impact}\\[4pt]
\includegraphics[width=0.80\columnwidth]{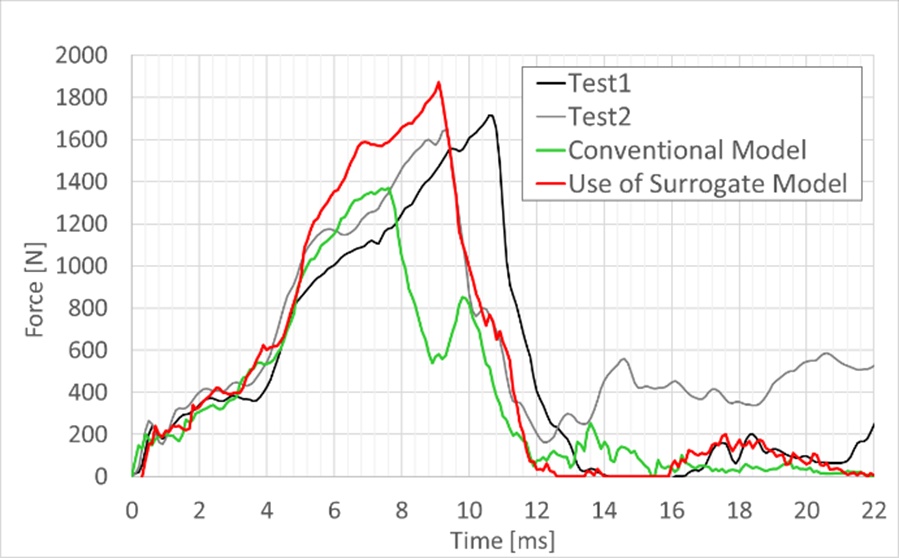}\\[1pt]
{\small (b) Impact force--time histories}
\caption{Practical impact on the target part. (a)~The isotropic model (left) over-spreads fracture; the surrogate-informed model (right) confines it to the test-observed locations. (b)~The surrogate-informed model (red) tracks the two test curves through the loading rise and peak; the isotropic baseline (green) collapses early.}
\label{fig:crash}
\end{figure}

The second effect is case-level crash accuracy, reported previously with the surrogate \citep{ito2026jsme} and summarized here as application evidence (Figure~\ref{fig:crash}). In test, bending under the impactor cracks the part at the two fixtures. The isotropic model (MAT\_215 replaced by an isotropic law, $\sim$2\,mm shell) under-predicted peak force by $19.0\%$ (1370 vs.\ 1682\,N test average) and fractured at spurious locations; with the predicted orientation field, peak force was 1873\,N---an $11.4\%$ absolute error, $7.6$ points better---and fracture stayed at the test-observed locations. The force--time history agrees: the isotropic baseline sheds load early, while the orientation-informed model tracks the measured rise, peak, and fracture timing. This connects the workflow to a real decision---whether a geometry passes pedestrian protection and where to reinforce it. Together, the two effects change the development sequence: gate layout, once fixed by molding feasibility alone, can now be screened jointly for moldability and crash performance.

\section{From Prototype to Production Tool}

\begin{figure}[t]
\centering
\includegraphics[width=0.96\columnwidth]{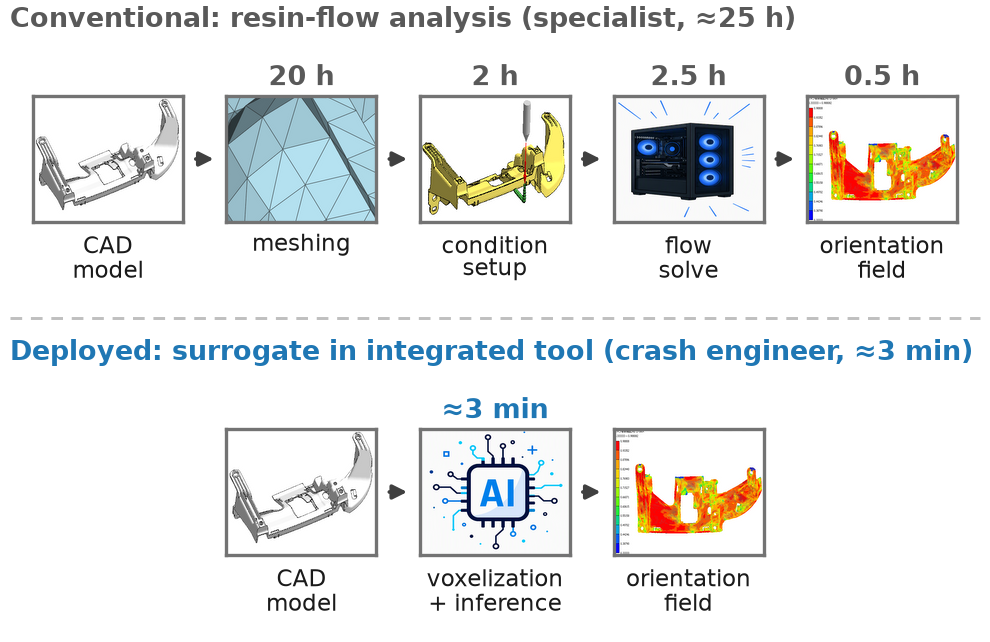}
\caption{Orientation-field preparation from the same CAD input to the same output. The conventional route chains meshing (20\,h), molding-condition setup (2\,h), flow solve (2.5\,h), and field measurement (0.5\,h)---about 25\,h of molding-specialist work; the deployed route obtains the field by voxelization and surrogate inference in about 3 minutes, and the full preparation including mapping, export, and overhead takes about 0.1\,h (Section~5). Gate prediction (about 10\,min, when gates are unknown) and the crash solve lie outside both chains.}
\label{fig:workflow}
\end{figure}

A validated model is not yet a tool. The engineers of Section~2 have no molding background, and at the early design stage even the injection gates are unknown---suppliers treat molding conditions as proprietary, and the production source is often undecided when crash analysis must run. Turning the surrogate into something those engineers use unaided required three further subsystems; the mapping and reporting subsystems were developed against an explicit \emph{zero-manual-steps} target. Each must do more than run unattended: it must hand the crash model exactly what the material card consumes. Figure~\ref{fig:workflow} contrasts the resulting preparation workflow with the conventional route.

\paragraph{Gate prediction from a molding database.} When no gate information exists, the system proposes it. From indices read off the CAD model---part volume, dimensions, flow ratio, flow length---a documented procedure built on the Section~3 survey, assisted by a commercial gate-position advisor, estimates gate number, type, and position. On held-out parts it predicts gate count with $90\%$ accuracy and position with a mean error of $11\%$ of flow length, in about 10 minutes. The prediction is an editable proposal---the engineer sets the gates manually in the GUI at the computed positions and adjusts them as needed---and, because the field regenerates in minutes, it is simply re-run when actual gate information becomes available. The procedure is captured in a 36-page document; as its operational test, an engineer from outside the team, with no molding background, completed a prediction unaided.

\begin{figure}[t]
\centering
\includegraphics[width=0.76\columnwidth]{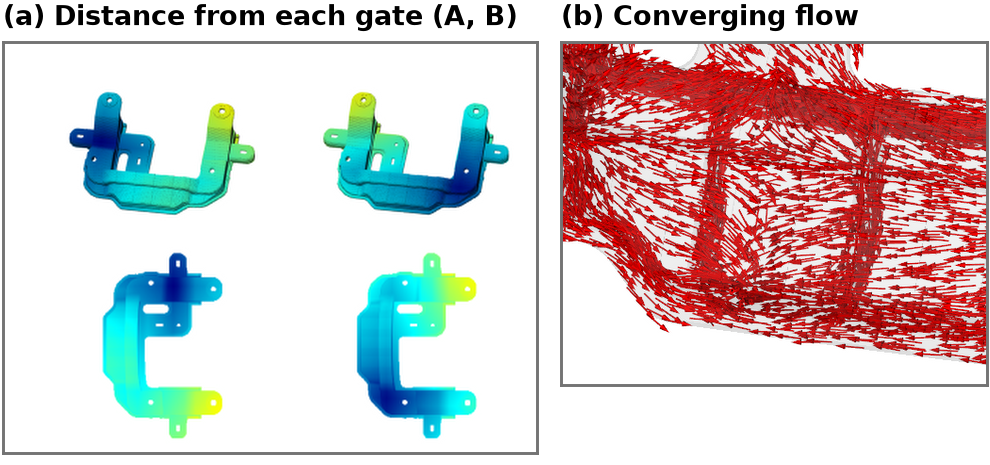}
\caption{Sample views from an automatically generated report: (a)~distance from each gate, shaded from dark blue at the gate to yellow farthest from it, and (b)~regions of converging flow. Together they let a reader anticipate likely weld-line locations without installing the tool.}
\label{fig:report}
\end{figure}

\paragraph{Automatic mapping into the crash model.} The predicted field must be written onto the crash model's shell mesh. Commercial mapping tools were rejected: they required manual preparation, took about 30 minutes per part, and could not write the proprietary in-house material card used alongside MAT\_215. A purpose-built mapper closed the gap with no manual steps, and successive optimization cut mapping time from 35 minutes in the first internal version to about 15 seconds. Both the standard and the proprietary material cards are supported through a single switch.

\paragraph{Through-thickness representation.} The crash shell model reads orientation at five through-thickness integration points---the outer points sampling the skin layers, the inner points the core; writing a single averaged direction there would discard exactly the anisotropy the material card is built to consume, so the mapper writes the surrogate's separately predicted core- and skin-layer fields to their corresponding points. At one position of the target bracket, for example, the flow-analysis cross-section shows a thin, highly aligned skin over a thicker, weakly aligned core, and the mapped integration points retain that distinction---flow-aligned skin fields at the outer points, a differently oriented core field at the mid-plane.
\paragraph{One GUI, with shareable reports.} The prediction, mapping, and reporting functions are driven from a single integrated GUI (Figure~\ref{fig:gui}); the conventional workflow required hopping across three tools and hand-carrying intermediate files. A routine run loads the STL, selects the model (including geometry-class modes such as the annular model of Section~4), sets the gates manually at the positions computed by the gate-prediction procedure, runs preprocessing and inference (about 3 minutes), inspects the field at hole and weld regions, then maps (15 seconds) and exports---with no manual file conversion between steps. A report generator captures selected views automatically, so results circulate to engineers who never installed the tool, and its gate-distance and flow-collision views let a reader infer likely weld-line locations from the report alone (Figure~\ref{fig:report}).

\section{Development Cost, Status, and Lessons Learned}

\paragraph{Timeline and cost.} The project was initiated in 2021; the controlled studies ran through 2021--2022; and the production model and tooling were built and passed internal stage reviews across 2023--2025. Every training pair requires one resin-flow solve---roughly 36{,}500 cases across the three stages---and training itself totaled roughly 1{,}450 hours (Table~\ref{tab:metrics}).

\paragraph{Governance before deployment.} Deployment was gated by an expert review with the pedestrian-protection CAE group. A first review (November 2024) made three demands---clarify the part and material scope, justify the $20\%$ load-error target and quantify the improvement, and verify other impact directions---whose responses were examined in a re-review (March 2025) that confirmed applicability and directed transfer to the development members. The gate mattered: applicability was confirmed by the people who own the safety decision.

\paragraph{Status and path to full deployment.} The system passed expert re-review in March 2025 and has been transferred to the crash-safety group; cumulative usage counts are confidential and still small. Table~\ref{tab:evidence} lays out the path set by the pedestrian-protection CAE group, from the completed stages to the next step toward full-vehicle use and the coverage expansion by paired-data regeneration and retraining. Maintenance follows the same economics, and the annular fix of Section~4 set the pattern: when validation exposes a weakness, a dedicated geometry-class model is added as a GUI mode.
\begin{table}[t]
\centering
\small
\setlength{\tabcolsep}{3pt}
\begin{tabular}{@{}>{\raggedright\arraybackslash}p{0.32\columnwidth}l>{\raggedright\arraybackslash}p{0.43\columnwidth}@{}}
\toprule
Stage & Status & Notes / next step \\
\midrule
Design-use approval & Completed & Re-review, Mar.\ 2025 \\
Transfer to crash-safety group & Completed & 2025 \\
Directed initial use & Under way & Post-test investigation, fracture-risk assessment, controlled-fracture design; counts confidential \\
Routine full-vehicle use & Pending & Next: a track record in the directed scenarios \\
Coverage expansion & Pending & New paired data and retraining for more material grades and gate counts \\
\bottomrule
\end{tabular}
\caption{Deployment status and path to full-scale use (non-confidential).}
\label{tab:evidence}
\end{table}

\paragraph{Effect of training data and representation.} Restricting training gates to manufacturable settings from the 46-part survey was intended to concentrate model capacity on cases the tool will actually meet; random gates would have filled training with configurations no mold designer would build. Likewise, the image representation that sufficed in 2D could not express flow climbing a rib, and switching to voxel distance fields---with no change to the GAN recipe---enabled the T-plate and production stages.

\paragraph{Staged validation and error reporting.} Each stage of the validation answered a different engineering objection---learnability at scale (2D), transfer to unseen 3D shapes (T-plates), performance on a real part (production holdouts)---and future parts can be judged by which stage they resemble. The gate-perturbation finding also shaped the reporting: a surrogate's output always looks smooth and plausible, even where it is wrong, so the system itself must point analysts to the regions where errors concentrate---holes and weld lines.

\paragraph{Role of shareable reports in adoption.} Beyond speed, the shareable report supported adoption: a prediction inside a tool reaches only the tool's users, whereas a circulating report reaches design engineers, managers, and suppliers, and its weld-line and gate-distance views give non-specialists something concrete to discuss.

\paragraph{Reuse in other domains.} The same development approach has already been reused by another division for aerodynamic prediction, and an aluminum-casting application is under study; only the data generator changed in the transfer.

\paragraph{Limitations.} The system covers one material family (PP-GF20) and at most two gates; more gates need new paired data and a new model. It predicts orientation only---weld-line strength reduction and warpage still come from the flow solver---and its acceptance metrics (field MSE, regional angular error) are proxies that bound the material-card error rather than a specific crash quantity---hence the component-level confirmation of Section~5. Gate prediction inherits the survey's scope: part classes outside it fall back to manual gate setting.

\section{Conclusion}

We developed a previously reported fiber-orientation surrogate into an engineering system for crash CAE at Honda: the orientation-preparation workflow drops from about 25 hours of molding-specialist work to about 0.1 hours that crash engineers run directly, with no manual file conversion from CAD to mapped LS-DYNA export and with layer-resolved fields written to the integration points. The system has passed expert review, been transferred to the crash-safety group, and entered initial, limited use in directed fracture-prediction scenarios. In the previously reported component validation, peak-force error fell from $19.0\%$ (isotropic baseline) to $11.4\%$ and predicted fracture stayed at the test-observed locations. The contribution is the transferable path from model to tool: constrained data, staged validation, expert review, and integrated tooling.

\bibliography{iaai_fiber_orientation}

\begin{thebibliography}{12}
\providecommand{\natexlab}[1]{#1}

\bibitem[{Fertschej, Reithofer, and Rollant(2016)}]{fertschej2016impetus}
Fertschej, A.; Reithofer, P.; and Rollant, M. 2016.
\newblock 4a Impetus (Part 2): Innovations -- Test Methods, {MAT\_SAMP-1},
  Anisotropy, Composites and More.
\newblock In \emph{Proceedings of the 14th German LS-DYNA Conference}.

\bibitem[{Guo, Li, and Iorio(2016)}]{guo2016flowcnn}
Guo, X.; Li, W.; and Iorio, F. 2016.
\newblock Convolutional Neural Networks for Steady Flow Approximation.
\newblock In \emph{Proceedings of the 22nd ACM SIGKDD International Conference
  on Knowledge Discovery and Data Mining}, 481--490.

\bibitem[{Hallquist(2006)}]{hallquist2006lsdyna}
Hallquist, J.~O. 2006.
\newblock \emph{{LS-DYNA} Theory Manual}.
\newblock Livermore Software Technology Corporation.

\bibitem[{{Humanetics}(2022)}]{humanetics2022apli}
{Humanetics}. 2022.
\newblock Advanced Pedestrian Legform Impactor ({aPLI}) Data Sheet.
\newblock \url{https://www.humaneticsgroup.com/}.
\newblock Total mass 25\,kg; simplified upper-body mass added to the 13\,kg
  FlexPLI.

\bibitem[{Isola et~al.(2017)Isola, Zhu, Zhou, and Efros}]{isola2017pix2pix}
Isola, P.; Zhu, J.-Y.; Zhou, T.; and Efros, A.~A. 2017.
\newblock Image-to-Image Translation with Conditional Adversarial Networks.
\newblock In \emph{Proceedings of the IEEE Conference on Computer Vision and
  Pattern Recognition}, 5967--5976.

\bibitem[{Ito and Nakagawa(2026)}]{ito2026jsme}
Ito, O.; and Nakagawa, Y. 2026.
\newblock Development of an Efficient Crash {CAE} Modelling Method Using a
  Fiber-Orientation Prediction Model.
\newblock \emph{Transactions of the JSME}, 92(957): 26--00041.
\newblock In Japanese.

\bibitem[{Nutini and Vitali(2010)}]{nutini2010anisotropy}
Nutini, M.; and Vitali, M. 2010.
\newblock Simulating Anisotropy with {LS-DYNA} in Glass-Reinforced,
  Polypropylene-Based Components.
\newblock In \emph{Proceedings of the German LS-DYNA Forum}.

\bibitem[{Nutini et~al.(2019)Nutini, Vitali, Benanti, and
  Formolo}]{nutini2019failure}
Nutini, M.; Vitali, M.; Benanti, M.; and Formolo, S. 2019.
\newblock Polypropylene Composites under Impact: Anisotropy, Mapping and
  Failure Criteria in Simulations, and Validation on a Part for Building and
  Construction Industry.
\newblock In \emph{Proceedings of the 12th European LS-DYNA Conference}.

\bibitem[{Ronneberger, Fischer, and Brox(2015)}]{ronneberger2015unet}
Ronneberger, O.; Fischer, P.; and Brox, T. 2015.
\newblock U-Net: Convolutional Networks for Biomedical Image Segmentation.
\newblock In \emph{Medical Image Computing and Computer-Assisted Intervention},
  234--241.

\bibitem[{Thuerey et~al.(2020)Thuerey, Wei{\ss}enow, Prantl, and
  Hu}]{thuerey2020airfoil}
Thuerey, N.; Wei{\ss}enow, K.; Prantl, L.; and Hu, X. 2020.
\newblock Deep Learning Methods for Reynolds-Averaged Navier--Stokes
  Simulations of Airfoil Flows.
\newblock \emph{AIAA Journal}, 58(1): 25--36.

\bibitem[{{Toray Engineering D Solutions Co., Ltd.}(2026)}]{timon2026}
{Toray Engineering D Solutions Co., Ltd.} 2026.
\newblock {3D TIMON} Resin Flow Analysis Software.
\newblock \url{https://www.3dtimon.com/}.
\newblock Accessed March 2026.

\bibitem[{Vinuesa and Brunton(2022)}]{vinuesa2022cfdml}
Vinuesa, R.; and Brunton, S.~L. 2022.
\newblock Enhancing Computational Fluid Dynamics with Machine Learning.
\newblock \emph{Nature Computational Science}, 2: 358--366.

\end{thebibliography}

\clearpage
\onecolumn
\appendix
\setcounter{figure}{0}
\setcounter{table}{0}
\renewcommand{\thefigure}{A\arabic{figure}}
\renewcommand{\thetable}{A\arabic{table}}
\section{Additional Detail}
\label{app:validation}

This appendix collects the network-architecture figure (Section~3), the staged-validation summary and figures (Section~4), and the tool screenshot (Section~6).

\begin{center}
\includegraphics[width=0.85\textwidth]{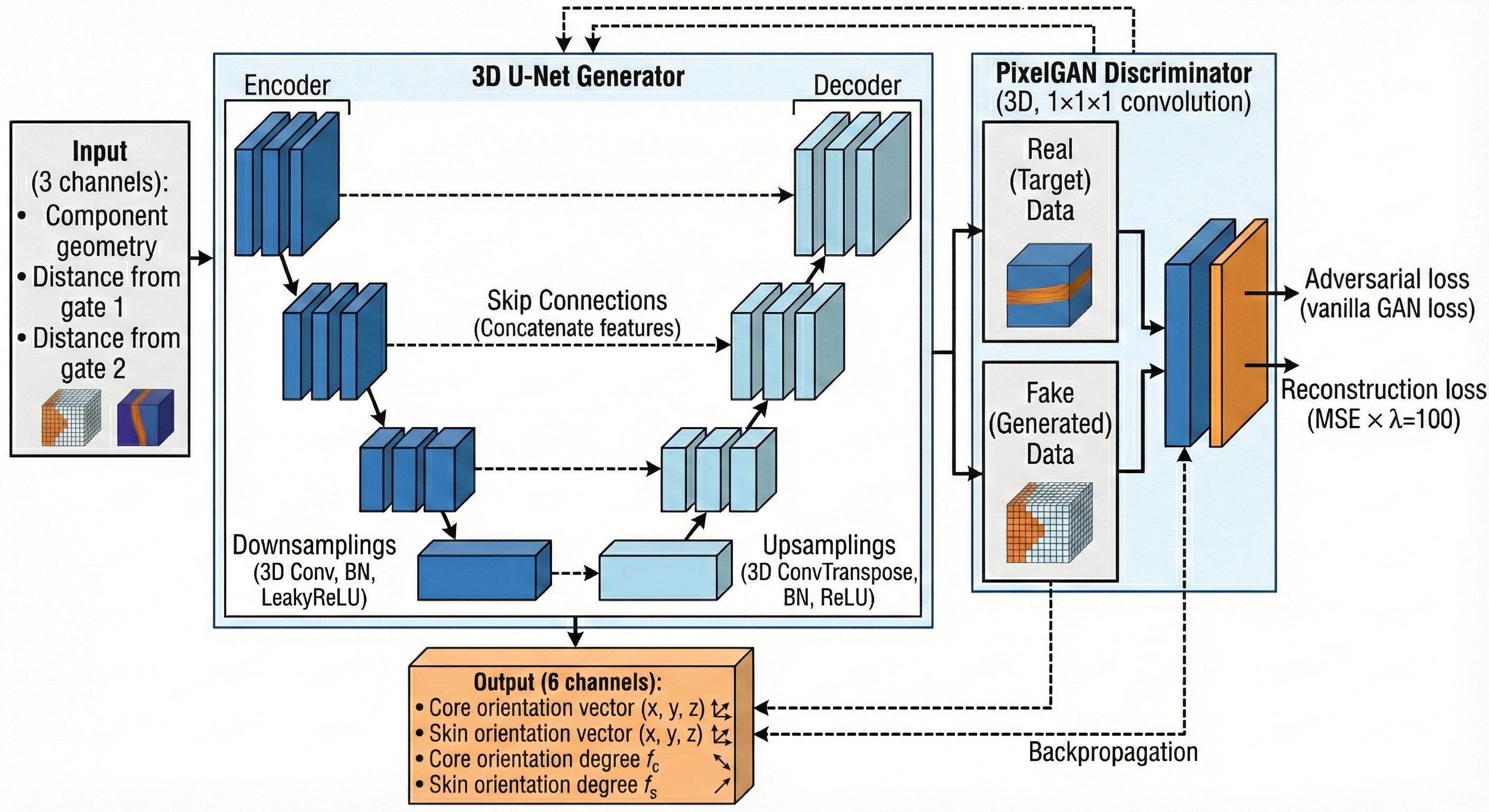}
\captionof{figure}{3D conditional GAN architecture. Gate-distance fields and geometry information are mapped to core- and skin-layer fiber orientation by a 3D U-Net generator with a local PatchGAN discriminator.}
\label{fig:network}
\end{center}

\begin{center}
\footnotesize
\setlength{\tabcolsep}{4pt}
\begin{tabular}{@{}p{0.13\textwidth}p{0.15\textwidth}p{0.24\textwidth}p{0.22\textwidth}p{0.11\textwidth}@{}}
\toprule
Stage & Training data (cases) & Evaluation target & Result & Training time \\
\midrule
Simple 2D & 22{,}044 & 4{,}845 unseen cases & MSE\,$\le$0.0175 & 120\,h \\
T-plate & 10{,}500 & 1{,}050 unseen shapes & MSE\,$\le$0.0248 & 331\,h \\
Actual part & 4{,}000 & one held-out part per gate count & 22--30$^\circ$ regional angular error & $\sim$1{,}000\,h \\
\bottomrule
\end{tabular}
\captionof{table}{Staged validation summary. Simple 2D and T-plate report MSE; actual part reports regional angular error.}
\label{tab:metrics}
\end{center}

\begin{center}
\includegraphics[width=0.96\textwidth]{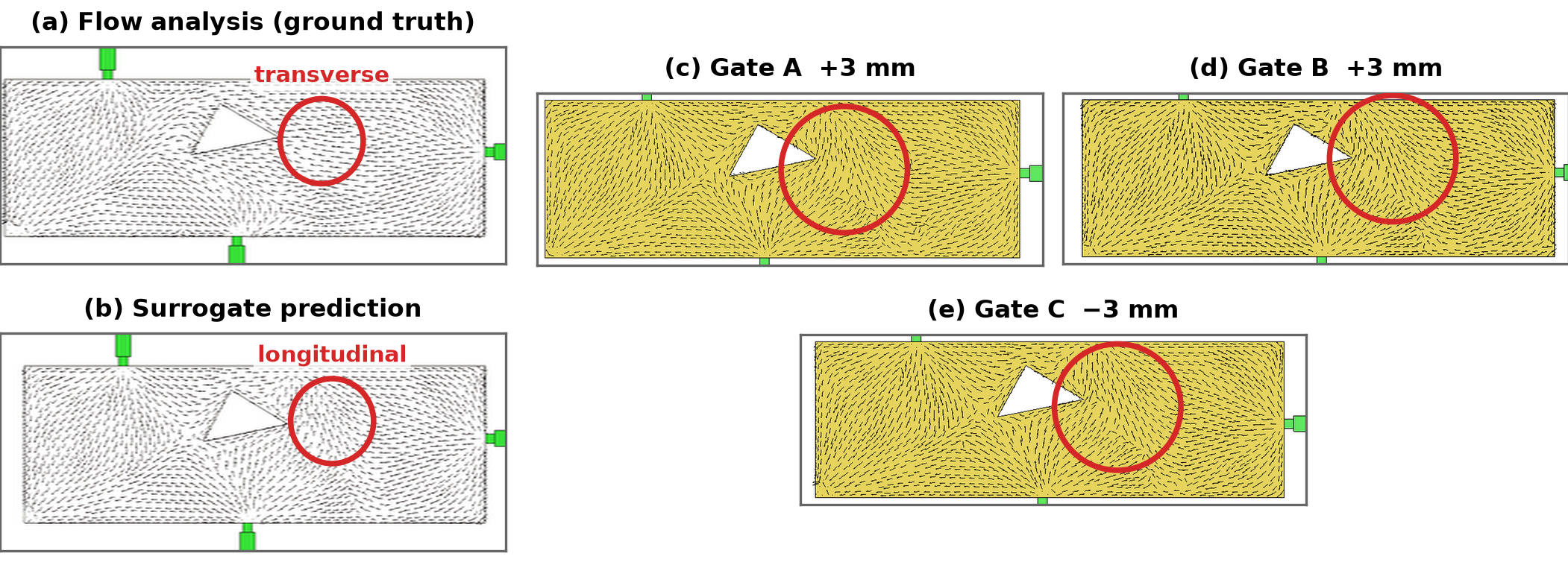}
\captionof{figure}{Gate-perturbation check on the maximum-error 2D case (Stage~1). The flow-analysis reference (a) is transverse in the circled region, whereas the surrogate (b) predicts longitudinal; additional flow-analysis runs with one gate displaced by 3\,mm (c--e) give longitudinal orientations there, and the original transverse orientation does not reappear.}
\label{fig:perturb}
\end{center}

\begin{center}
\includegraphics[width=0.60\textwidth]{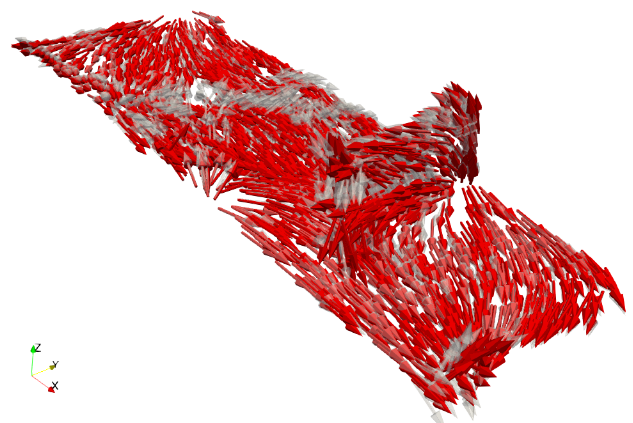}\\[2pt]
{\small\predgtlegend}
\captionof{figure}{Stage 2 representative average-error case (MSE 0.0029): surrogate prediction over the flow-analysis reference. Across the 1{,}050 unseen-shape holdouts, MSE ranged from 0.0004 to 0.0248.}
\label{fig:tplate}
\end{center}

\begin{center}
\includegraphics[width=0.55\textwidth]{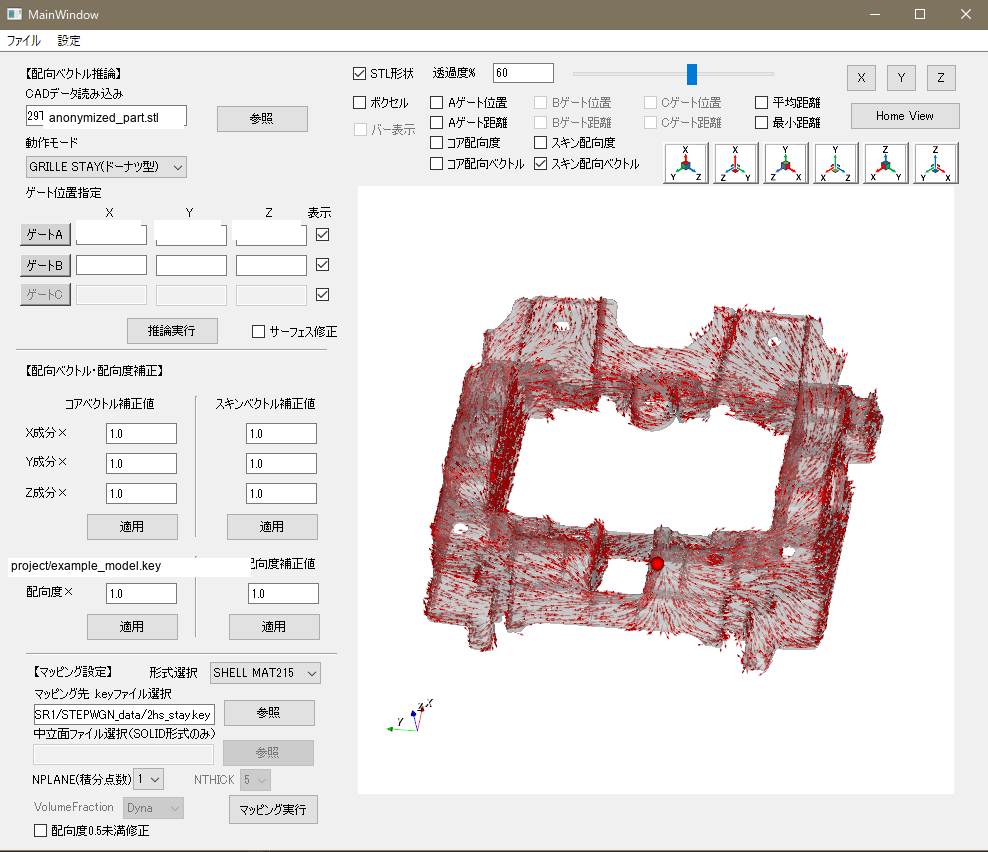}
\captionof{figure}{The integrated tool: a single GUI drives gate handling, voxelization, prediction, mapping, and report generation; the panel shows a predicted orientation field on a production part.}
\label{fig:gui}
\end{center}

\section{Injection-Molding Background: Gates and Skin/Core Layers}
\label{app:molding}

This appendix provides background on two injection-molding terms used throughout the paper, for readers unfamiliar with plastic-part manufacturing.

\paragraph{Gates.} A \emph{gate} is the opening in the injection mold through which molten resin enters the cavity. Two gate types dominate automotive practice. A \emph{side gate} feeds the part from an edge; it can be placed almost anywhere along the part periphery, but its higher flow resistance makes it unsuitable for large parts. A \emph{direct gate} injects onto a part surface; it sustains higher injection pressure and therefore suits large parts, but it requires a flat area large enough to seat it. The number and placement of gates determine the flow paths inside the cavity---and hence where flow fronts meet to form weld lines and how fibers align along the way. This is why the gate layout is an input to both the flow solver and the surrogate (Section~3), and why the system provides gate-setting support for the early design stage, when suppliers have not yet disclosed the gate information (Section~6).

\paragraph{Skin and core layers.} In an injection-molded short-fiber part, fiber orientation varies through the wall thickness. During filling, the melt near the mold walls undergoes strong shear, so fibers in the near-surface \emph{skin} layers align closely with the local flow direction; around mid-thickness the shear is weak, and fibers in the \emph{core} layer are less strongly aligned and can point in a different direction. Because the two layers form under different shear histories, the flow solver reports their in-plane orientations separately, and the surrogate predicts them separately as its six output channels: an orientation vector and an orientation degree for each layer (Section~3). The distinction matters mechanically because the near-surface skin layers dominate the bending and crash response of thin-walled parts: a single thickness-averaged direction would misstate exactly the anisotropy the crash material card is built to capture, which is why the mapper writes the two layers to their corresponding through-thickness integration points (Section~6).

\end{document}